\documentclass[11pt,letterpaper]{article}
\usepackage{emnlp2017}
\usepackage{times}
\usepackage{latexsym}
\usepackage{graphicx}
\usepackage{multirow}
\usepackage{verbatim}
\usepackage{amsmath}
\usepackage{subcaption}
\usepackage{array}
\newcolumntype{P}[1]{>{\centering\arraybackslash}p{#1}}
\newcolumntype{M}[1]{>{\centering\arraybackslash}m{#1}}

\emnlpfinalcopy

\def\emnlppaperid{871}

\title{Supervised Learning of Universal Sentence Representations from\\ Natural Language Inference Data}

\author{
Alexis Conneau \\
Facebook AI Research \\
\texttt{aconneau@fb.com} \\
\And
Douwe Kiela \\
Facebook AI Research \\
\texttt{dkiela@fb.com} \\
\And
Holger Schwenk \\
Facebook AI Research \\
\texttt{schwenk@fb.com} \\
\AND
Lo\"{\i}c Barrault \\
LIUM, Universit\'e Le Mans \\
\texttt{loic.barrault@univ-lemans.fr} \\
\And
Antoine Bordes \\
Facebook AI Research \\
\texttt{abordes@fb.com} \\
}

\date{}

\begin{document}

\maketitle

\begin{abstract}
Many modern NLP systems rely on word embeddings, previously trained in an unsupervised manner on large corpora, as base features.
Efforts to obtain embeddings for larger chunks of text, such as sentences, have however not been so successful.
Several attempts at learning unsupervised representations of sentences have not reached satisfactory enough performance to be widely adopted.
In this paper, we show how universal sentence representations trained using the supervised data of the Stanford Natural Language Inference datasets can consistently outperform unsupervised methods like SkipThought vectors \cite{kiros2015skip} on a wide range of transfer tasks. Much like how computer vision uses ImageNet to obtain features, which can then be transferred to other tasks, our work tends to indicate the suitability of natural language inference for transfer learning to other NLP tasks. Our encoder is publicly available\footnote{\url{https://www.github.com/facebookresearch/InferSent}}.

\end{abstract}

\section{Introduction}

Distributed representations of words (or word embeddings) \cite{bengio2003neural, collobert2011natural, mikolov2013distributed, pennington2014glove} have shown to provide useful features for various tasks in natural language processing and computer vision. 
While there seems to be a consensus concerning the usefulness of word embeddings and how to learn them, this is not yet clear with regard to representations that carry the meaning of a full sentence. That is, how to capture the relationships among multiple words and phrases in a single vector remains an question to be solved.

In this paper, we study the task of learning universal representations of sentences, i.e., a sentence encoder model that is trained on a large corpus and subsequently transferred to other tasks.
Two questions need to be solved in order to build such an encoder, namely: what is the preferable neural network architecture; and how and on what task should such a network be trained.
Following existing work on learning word embeddings, most current approaches consider learning sentence encoders in an unsupervised manner like SkipThought \cite{kiros2015skip} or FastSent \cite{hill2016learning}. Here, we investigate whether supervised learning can be leveraged instead, taking inspiration from previous results in computer vision, where many models are pretrained on the ImageNet \cite{deng2009imagenet} before being transferred.
We compare sentence embeddings trained on various supervised tasks, and show that sentence embeddings generated from models trained on a natural language inference (NLI) task reach the best results in terms of transfer accuracy.
We hypothesize that the suitability of NLI as a training task is caused by the fact that it is a high-level understanding task that involves reasoning about the semantic relationships within sentences.

Unlike in computer vision, where convolutional neural networks are predominant, there are multiple ways to encode a sentence using neural networks. Hence, we investigate the impact of the sentence encoding architecture on representational transferability, and compare convolutional, recurrent and even simpler word composition schemes.
Our experiments show that an encoder based on a bi-directional LSTM architecture with max pooling, trained on the Stanford Natural Language Inference (SNLI) dataset \cite{bowman2015large}, yields state-of-the-art sentence embeddings compared to all existing alternative unsupervised approaches like SkipThought or FastSent, while being much faster to train. We establish this finding on a broad and diverse set of transfer tasks that measures the ability of sentence representations to capture general and useful information.

\section{Related work}
Transfer learning using supervised features has been successful in several computer vision applications \cite{razavian2014cvpr}. Striking examples include face recognition \cite{taigman2014deepface} and visual question answering \cite{antol2015vqa}, where image features trained on ImageNet \cite{deng2009imagenet} and word embeddings trained on large unsupervised corpora are combined. 

In contrast, most approaches for sentence representation learning are unsupervised, arguably because the NLP community has not yet found the best supervised task for embedding the semantics of a whole sentence. Another reason is that neural networks are very good at capturing the biases of the task on which they are trained, but can easily forget the overall information or semantics of the input data by specializing too much on these biases. Learning models on large unsupervised task makes it harder for the model to specialize. \newcite{littwin2016multiverse} showed that co-adaptation of encoders and classifiers, when trained end-to-end, can negatively impact the generalization power of image features generated by an encoder. They propose a loss that incorporates multiple orthogonal classifiers to counteract this effect.

Recent work on generating sentence embeddings range from models that compose word embeddings \cite{le2014distributed, arora2016asimple, wieting2015towards} to more complex neural network architectures. SkipThought vectors \cite{kiros2015skip} propose an objective function that adapts the skip-gram  model for words \cite{mikolov2013distributed} to the sentence level. By encoding a sentence to predict the sentences around it, and using the features in a linear model, they were able to demonstrate good performance on 8 transfer tasks. They further obtained better results using layer-norm regularization of their model in \cite{ba2016layer}. \newcite{hill2016learning} showed that the task on which sentence embeddings are trained significantly impacts their quality.

In addition to unsupervised methods, they included supervised training in their comparison---namely, on machine translation data (using the WMT'14 English/French and English/German pairs), dictionary definitions and image captioning data (see also \citet{kiela2017learning}) from the COCO dataset \cite{lin2014microsoft}. These models obtained significantly lower results compared to the unsupervised Skip-Thought approach.

Recent work has explored training sentence encoders on the SNLI corpus and applying them on the SICK corpus \cite{marelli2014sick}, either using multi-task learning or pretraining \cite{mou2016transferable, bowman2015large}. The results were inconclusive and did not reach the same level as simpler approaches that directly learn a classifier on top of unsupervised sentence embeddings instead \cite{arora2016asimple}. To our knowledge, this work is the first attempt to fully exploit the SNLI corpus for building generic sentence encoders. As we show in our experiments, we are able to consistently outperform unsupervised approaches, even if our models are trained on much less (but human-annotated) data.

\section{Approach}

This work combines two research directions, which we describe in what follows.
First, we explain how the NLI task can be used to train universal sentence encoding models using the SNLI task.
We subsequently describe the architectures that we investigated for the sentence encoder, which, in our opinion, covers a suitable range of sentence encoders currently in use. Specifically, we examine standard recurrent models such as LSTMs and GRUs, for which we investigate mean and max-pooling over the hidden representations; a self-attentive network that incorporates different views of the sentence; and a hierarchical convolutional network that can be seen as a tree-based method that blends different levels of abstraction. 

\subsection{The Natural Language Inference task}
The SNLI dataset consists of 570k human-generated English sentence pairs, manually labeled with one of three categories: entailment, contradiction and neutral. It captures natural language inference, also known in previous incarnations as Recognizing Textual Entailment (RTE), and constitutes one of the largest high-quality labeled resources explicitly constructed in order to require understanding sentence semantics.
We hypothesize that the semantic nature of NLI makes it a good candidate for learning universal sentence embeddings in a supervised way. That is, we aim to demonstrate that sentence encoders trained on natural language inference are able to learn sentence representations that capture universally useful features.

\begin{figure}[ht]
  \centering
  \includegraphics[width=0.78\linewidth]{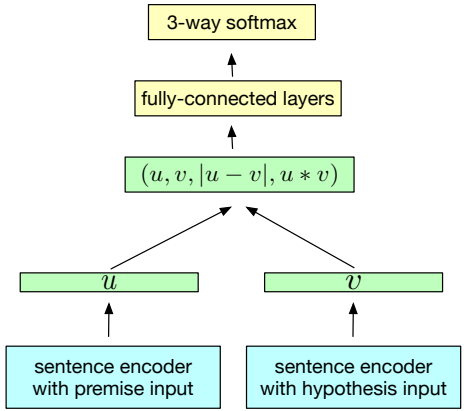}
    \caption{\bf Generic NLI training scheme.}
    \label{fig:snli}
\end{figure}

Models can be trained on SNLI in two different ways: (i) sentence encoding-based models that explicitly separate the encoding of the individual sentences and (ii) joint methods that allow to use encoding of both sentences (to use cross-features or attention from one sentence to the other).

Since our goal is to train a generic sentence encoder, we adopt the first setting. As illustrated in Figure~\ref{fig:snli}, a typical architecture of this kind uses a shared sentence encoder that outputs a representation for the premise $u$ and the hypothesis $v$. Once the sentence vectors are generated, 3 matching methods are applied to extract relations between $u$ and $v$ : (i) concatenation of the two representations $(u,v)$; (ii) element-wise product $u*v$; and (iii) absolute element-wise difference $\vert u-v\vert$. The resulting vector, which captures information from both the premise and the hypothesis, is fed into a 3-class classifier consisting of multiple fully-connected layers culminating in a softmax layer.

\subsection{Sentence encoder architectures}
A wide variety of neural networks for encoding sentences into fixed-size representations exists, and it is not yet clear which one best captures generically useful information. We compare 7 different architectures: standard recurrent encoders with either Long Short-Term Memory (LSTM) or Gated Recurrent Units (GRU), concatenation of last hidden states of forward and backward GRU, Bi-directional LSTMs (BiLSTM) with either mean or max pooling, self-attentive network and hierarchical convolutional networks.

\subsubsection{LSTM and GRU}
Our first, and simplest, encoders apply recurrent neural networks using either LSTM \cite{hochreiter1997long} or GRU \cite{cho2014properties} modules, as in sequence to sequence encoders \cite{sutskever2014sequence}.
For a sequence of $T$ words $(w_1, \ldots, w_T)$, the network computes a set of $T$ hidden representations $h_1, \ldots, h_T$, 
with $h_t\,=\,\overrightarrow{\text{LSTM}}(w_1, \ldots, w_T)$ (or using GRU units instead). A sentence is represented by the last hidden vector, $h_T$.

We also consider a model BiGRU-last that concatenates the last hidden state of a forward GRU, and the last hidden state of a backward GRU to have the same architecture as for SkipThought vectors.

\subsubsection{BiLSTM with mean/max pooling}
For a sequence of T words $\{ w_{t}\}_{t=1,\ldots,T}$, a bidirectional LSTM computes a set of T vectors $\{h_t\}_t$. For $t\in [1,\ldots, T]$, $h_t$, is the concatenation of a forward LSTM and a backward LSTM that read the sentences in two opposite directions: 
\begin{eqnarray*}
\overrightarrow{h_t} &=& \overrightarrow{\text{LSTM}}_t( w_1, \ldots, w_T) \\
\overleftarrow{h_t} &=& \overleftarrow{\text{LSTM}}_t(w_1, \ldots, w_T) \\
h_t &=& [\overrightarrow{h_t}, \overleftarrow{h_t}]
\end{eqnarray*}
We experiment with two ways of combining the varying number of $\{h_t\}_t$ to form a fixed-size vector, either by selecting the maximum value over each dimension of the hidden units (max pooling) \cite{collobert2008unified} or by considering the average of the representations (mean pooling).

\begin{figure}[h!]
  \centering
  \includegraphics[width=0.56\linewidth]{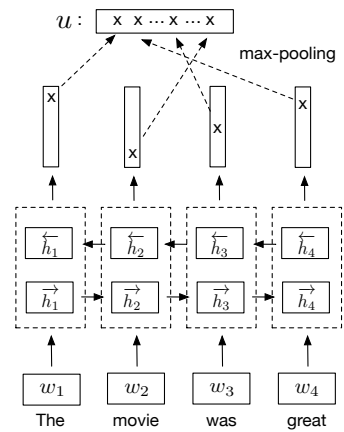}
    \caption{\bf Bi-LSTM max-pooling network.}
  \label{fig:blstmmax}
\end{figure}

\subsubsection{Self-attentive network}
The self-attentive sentence encoder \cite{liu2016learning, lin2017structured} uses an attention mechanism over the hidden states of a BiLSTM to generate a representation $u$ of an input sentence. The attention mechanism is defined as :
\vspace{-1pt}
\begin{eqnarray*}
\bar{h}_i & = & \tanh(Wh_i + b_w) \\
\alpha_i &=& \frac{e^{\bar{h}_i^T u_w}}{\sum_i e^{\bar{h}_i^T u_w}} \\
u &=& \sum_t \alpha_i h_i 
\end{eqnarray*}

where $\{h_1,\ldots,h_T\}$ are the output hidden vectors of a BiLSTM. These are fed to an affine transformation ($W$, $b_w$) which outputs a set of keys $(\bar{h}_1,\ldots, \bar{h}_T)$. The $\{\alpha_i\}$ represent the score of similarity between the keys and a learned context query vector $u_w$. These weights are used to produce the final representation $u$, which is a weighted linear combination of the hidden vectors.

\begin{figure}[h!]
  \centering
  \includegraphics[width=0.75\linewidth]{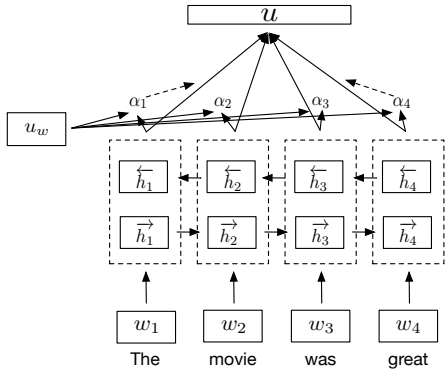}
    \caption{\bf Inner Attention network architecture.}
  \label{fig:inneratt}
\end{figure}

Following \citet{lin2017structured} we use a self-attentive network with multiple views of the input sentence, so that the model can learn which part of the sentence is important for the given task. Concretely, we have 4 context vectors $u^1_w, u^2_w, u^3_w, u^4_w$ which generate 4 representations that are then concatenated to obtain the sentence representation $u$. Figure~\ref{fig:inneratt} illustrates this architecture.

\subsubsection{Hierarchical ConvNet}
\begin{figure}[h!]
  \centering
  \includegraphics[width=0.9\linewidth]{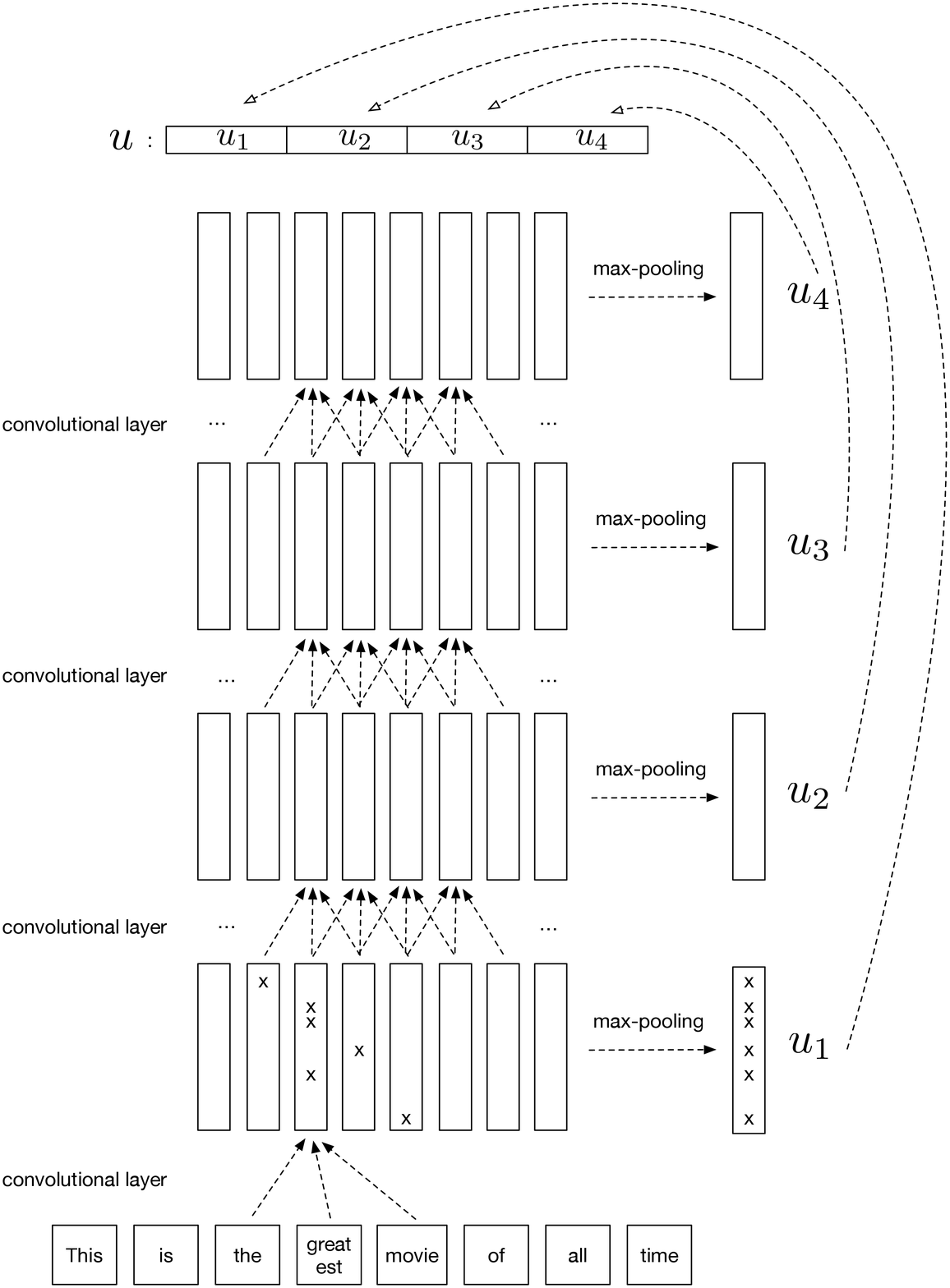}
    \caption{\bf Hierarchical ConvNet architecture.}
    \label{fig:hconvnet}
\end{figure}

One of the currently best performing models on classification tasks is a convolutional architecture termed {\it AdaSent} \cite{zhao2015self}, which concatenates different representations of the sentences at different level of abstractions. Inspired by this architecture, we introduce a faster version consisting of 4 convolutional layers. At every layer, a representation $u_i$ is computed by a max-pooling operation over the feature maps (see Figure~\ref{fig:hconvnet}).

The final representation $u = [u_1, u_2, u_3, u_4]$ concatenates representations at different levels of the input sentence. The model thus captures hierarchical abstractions of an input sentence in a fixed-size representation.

\subsection{Training details}
For all our models trained on SNLI, we use SGD with a learning rate of 0.1 and a weight decay of 0.99. At each epoch, we divide the learning rate by 5 if the dev accuracy decreases. We use mini-batches of size 64 and training is stopped when the learning rate goes under the threshold of $10^{-5}$. For the classifier, we use a multi-layer perceptron with 1 hidden-layer of 512 hidden units. We use open-source GloVe vectors trained on Common Crawl 840B with 300 dimensions as fixed word embeddings.

\begin{table*}[h!]
{\small
\centering
\begin{tabular}{|l|r|l|c|p{9.2cm}|}
\hline \bf name & \bf N & \bf task & \bf C & \bf examples \\
\hline
MR & 11k &  sentiment (movies) & 2 &  "Too slow for a younger crowd , too shallow for an older one." (neg)\\
CR & 4k &  product reviews & 2 & "We tried it out christmas night and it worked great ." (pos)\\
SUBJ & 10k & subjectivity/objectivity & 2 &  "A movie that doesn't aim too high , but doesn't need to." (subj) \\
MPQA & 11k &  opinion polarity & 2 & "don't want";  "would like to tell"; (neg, pos)\\
TREC & 6k & question-type & 6 &  "What are the twin cities ?" (LOC:city)\\
SST & 70k &  sentiment (movies) &  2 & "Audrey Tautou has a knack for picking roles that magnify her [..]" (pos) \\
\hline
\end{tabular}
\caption{\label{table:classif} {\bf Classification tasks}. C is the number of class and N is the number of samples.}

}%
\end{table*}

\section{Evaluation of sentence representations}
Our aim is to obtain general-purpose sentence embeddings that capture generic information that is useful for a broad set of tasks. To evaluate the quality of these representations, we use them as features in 12 transfer tasks. We present our sentence-embedding evaluation procedure in this section.
We constructed a sentence evaluation tool\footnote{\url{https://www.github.com/facebookresearch/SentEval}} called \textit{SentEval} \cite{conneau2018senteval} to automate evaluation on all the tasks mentioned in this paper. The tool uses Adam \cite{kingma2014adam} to fit a logistic regression classifier, with batch size 64.

\paragraph{Binary and multi-class classification}
We use a set of binary classification tasks (see Table~\ref{table:classif}) that covers various types of sentence classification, including sentiment analysis (MR, SST), question-type (TREC), product reviews (CR), subjectivity/objectivity (SUBJ) and opinion polarity (MPQA). We generate sentence vectors and train a logistic regression on top. A linear classifier requires fewer parameters than an MLP and is thus suitable for small datasets, where transfer learning is especially well-suited. We tune the L2 penalty of the logistic regression with grid-search on the validation set.

\paragraph{Entailment and semantic relatedness}
We also evaluate on the SICK dataset for both entailment (SICK-E) and semantic relatedness (SICK-R). We use the same matching methods as in SNLI and learn a Logistic Regression on top of the joint representation. 
For semantic relatedness evaluation, we follow the approach of \cite{tai2015improved} and learn to predict the probability distribution of relatedness scores.
We report Pearson correlation. 

\paragraph{STS14 - Semantic Textual Similarity}
While semantic relatedness is supervised in the case of SICK-R, we also evaluate our embeddings on the 6 unsupervised SemEval tasks of STS14 \cite{agirre-EtAl:2014:SemEval}. This dataset includes subsets of news articles, forum discussions, image descriptions and headlines from news articles containing pairs of sentences (lower-cased), labeled with a similarity score between 0 and 5. These tasks evaluate how the cosine distance between two sentences correlate with a human-labeled similarity score through Pearson and Spearman correlations.
\begin{table*}[h!]
{\small
\centering
\begin{tabular}{|l|l|r|p{4.6cm}|p{4.6cm}|c|}
\hline \bf name & \bf task & \bf N & \bf premise & \bf hypothesis & \bf label \\
\hline
SNLI & NLI & 560k & "Two women are embracing while holding to go packages." & "Two woman are holding packages." & entailment\\ 
SICK-E & NLI & 10k & A man is typing on a machine used for stenography & The man isn't operating a stenograph & contradiction\\
SICK-R & STS & 10k & "A man is singing a song and playing the guitar" & "A man is opening a package that contains headphones" & 1.6\\
STS14 & STS & 4.5k & "Liquid ammonia leak kills 15 in Shanghai" & "Liquid ammonia leak kills at least 15 in Shanghai" & 4.6\\
\hline
\end{tabular}
\caption{\label{table:sts} {\bf Natural Language Inference and Semantic Textual Similarity tasks}. NLI labels are contradiction, neutral and entailment. STS labels are scores between 0 and 5.}
}%
\end{table*}

\paragraph{Paraphrase detection}
The Microsoft Research Paraphrase Corpus is composed of pairs of sentences which have been extracted from news sources on the Web. Sentence pairs have been human-annotated according to whether they capture a paraphrase/semantic equivalence relationship. We use the same approach as with SICK-E, except that our classifier has only 2 classes.

\paragraph{Caption-Image retrieval}
The caption-image retrieval task evaluates joint image and language feature models \cite{hodosh2013flickr,lin2014microsoft}. The goal is either to rank a large collection of images by their relevance with respect to a given query caption (Image Retrieval), or ranking captions by their relevance for a given query image (Caption Retrieval). We use a pairwise ranking-loss $\mathcal{L}_{\text{cir}}(x,y)$:

\begin{align*}
\sum_y \sum_k \max (0, \alpha - s(Vy, Ux) + s(Vy, Ux_k))\,+\\
\sum_x \sum_{k'} \max (0, \alpha - s(Ux, Vy) + s(Ux, Vy_{k'}))
\end{align*}

where $(x,y)$ consists of an image $y$ with one of its associated captions $x$, $(y_{k})_{k}$ and $(y_{k'})_{k'}$ are negative examples of the ranking loss, $\alpha$ is the margin and $s$ corresponds to the cosine similarity. $U$ and $V$ are learned linear transformations that project the caption $x$ and the image $y$ to the same embedding space. We use a margin $\alpha=0.2$ and $30$ contrastive terms. We use the same splits as in \cite{karpathy2015deep}, i.e., we use 113k images from the COCO dataset (each containing 5 captions) for training, 5k images for validation and 5k images for test. For evaluation, we split the 5k images in 5 random sets of 1k images on which we compute Recall@K, with K $\in \{1,5,10\}$ and median (Med r) over the 5 splits. For fair comparison, we also report SkipThought results in our setting, using 2048-dimensional pretrained ResNet-101 \cite{he2016resnet} with 113k training images.

\begin{table}[h!]
{
\small
\centering
  \begin{tabular}{l|c|cc|cc}
    \multirow{2}{*}{\bf Model} &
      \multicolumn{1}{c|}{} &
      \multicolumn{2}{c|}{\bf NLI} &
      \multicolumn{2}{c}{\bf Transfer}\\
     & \bf dim & \bf dev & \bf test & \bf micro & \bf macro \\
    \hline
    LSTM & 2048 & 81.9 & 80.7 & 79.5 & 78.6 \\
    GRU & 4096 & 82.4 & 81.8 & 81.7 & 80.9 \\
    BiGRU-last & 4096 & 81.3 & 80.9 & 82.9 & 81.7 \\
    BiLSTM-Mean & 4096 & 79.0 & 78.2 & 83.1 & 81.7 \\
    Inner-attention & 4096 &  82.3 & 82.5 & 82.1 & 81.0 \\
    HConvNet & 4096 &  83.7 & 83.4 & 82.0 & 80.9 \\
    BiLSTM-Max & 4096 & \bf 85.0 & \bf \underline{84.5} & \bf 85.2 & \bf 83.7 \\

  \end{tabular}
\caption{\label{table:archi_results} {\bf Performance of sentence encoder architectures} on SNLI and (aggregated) transfer tasks. Dimensions of embeddings were selected according to best aggregated scores (see Figure \ref{fig:embed_results}).}

}%
\end{table}

\begin{table*}[h!]
\resizebox{1\linewidth}{!}{
\begin{tabular}{@{}l@{\,}|cccc@{\,\,}c@{\,\,}ccccc@{}}
\hline  \bf Model & \bf MR & \bf CR & \bf SUBJ & \bf MPQA & \bf SST & \bf TREC & \bf MRPC & \bf SICK-R & \bf SICK-E & \bf STS14 \\
\hline
\hline
\multicolumn{7}{l}{\it Unsupervised representation training (unordered sentences)} &&&& \\
\hline
Unigram-TFIDF & 73.7 & \bf 79.2 & 90.3 & 82.4 & - & 85.0 & 73.6/81.7 & - & - & .58/.57 \\
ParagraphVec (DBOW) & 60.2 & 66.9 & 76.3 & 70.7 & - & 59.4 & 72.9/81.1 & - & - & .42/.43 \\
SDAE & 74.6 & 78.0 & 90.8 &  86.9 & - & 78.4 & \textbf{73.7}/80.7 & - & - & .37/.38 \\
SIF (GloVe + WR)& - & - & - & - & 82.2 & - & - & - & \bf 84.6 & \textbf{.69}/ - \\
\hline
word2vec BOW$^{\dagger}$ & 77.7 & 79.8 & 90.9 & 88.3 & 79.7 & 83.6 & 72.5/81.4 & 0.803 & 78.7 & .65/.64 \\
fastText BOW$^{\dagger}$ & 78.3 & 81.0 & \bf 92.4 & 87.8 & \bf 81.9 & 84.8 & \bf 73.9/82.0 & 0.815 & 78.3 & .63/.62\\ 
GloVe BOW$^{\dagger}$ & \bf 78.7 & 78.5 & 91.6 & 87.6 & 79.8 & 83.6 & 72.1/80.9 & 0.800 & 78.6 & .54/.56 \\
GloVe Positional Encoding$^{\dagger}$ & 78.3 & 77.4 & 91.1 & 87.1 & 80.6 & 83.3 &  72.5/81.2 & 0.799 & 77.9 & .51/.54 \\

BiLSTM-Max (untrained)$^{\dagger}$  & 77.5 & \bf 81.3 & 89.6 & \bf 88.7 & 80.7 & \bf 85.8 & 73.2/81.6 & \bf 0.860 & 83.4 & .39/.48 \\
\hline
\hline
\multicolumn{7}{l}{\it Unsupervised representation training (ordered sentences)} &&&& \\
\hline
FastSent & 70.8 & 78.4 & 88.7 & 80.6 & - & 76.8 & 72.2/80.3 & - & - & \bf .63/.64 \\
FastSent+AE & 71.8 & 76.7 & 88.8 & 81.5 & - & 80.4 & 71.2/79.1 & - & - & .62/.62 \\
SkipThought &   76.5 & 80.1 & 93.6 & 87.1 & 82.0 & \bf \underline{92.2} & \bf 73.0/82.0 & \bf 0.858 & 82.3 & .29/.35 \\
SkipThought-LN & \bf 79.4 & \bf 83.1 & \bf \underline{93.7} & \bf 89.3 & 82.9 & 88.4 & - & \bf 0.858 & 79.5 & .44/.45 \\
\hline
\hline
\multicolumn{5}{l}{\it Supervised representation training} &&&&&& \\
\hline
CaptionRep (bow) & 61.9 &  69.3 & 77.4 &  70.8 &  - & 72.2 & 73.6/81.9 & - &  - & .46/.42 \\
DictRep (bow)  & 76.7 & 78.7 & 90.7 & 87.2 & - & 81.0 & 68.4/76.8 & - & - &  \bf .67/\underline{.70} \\
NMT En-to-Fr &  64.7 &  70.1 & 84.9 &  81.5 &  - & 82.8 & 69.1/77.1 & - &   & .43/.42 \\
Paragram-phrase  & - & - & - & - & 79.7 & - & - & 0.849 & 83.1 & \multicolumn{1}{c}{\underline{\textbf{.71}}/ -} \\
\hline
BiLSTM-Max (on SST)$^{\dagger}$ & (*) & 83.7 & 90.2 & 89.5 & (*) & 86.0 & 72.7/80.9 & 0.863 & 83.1 & .55/.54  \\
BiLSTM-Max (on SNLI)$^{\dagger}$ &  79.9 & 84.6 & 92.1 & \bf 89.8 & 83.3 & \bf 88.7 & 75.1/82.3 & \bf \underline{0.885} & \bf \underline{86.3} & .68/.65 \\
BiLSTM-Max (on AllNLI)$^{\dagger}$ &  \bf \underline{81.1} & \bf \underline{86.3} & \bf 92.4 & \bf \underline{90.2} & \bf \underline{84.6} & 88.2 & \bf \underline{76.2}/\underline{83.1} & \bf \underline{0.884} & \bf \underline{86.3} & \bf .70/.67 \\
\hline
\hline
\multicolumn{7}{l}{\it Supervised methods (directly trained for each task -- no transfer)} && \\
\hline
Naive Bayes - SVM  & 79.4 & 81.8 & 93.2 & 86.3 &  83.1 & - & - & - & - & - \\
AdaSent  &  83.1 & 86.3 & 95.5 & 93.3 & - & 92.4 & - & - & - & - \\
TF-KLD  &  - & - & - & - & - & - & 80.4/85.9 & - & - & - \\
Illinois-LH  &  - & - & - & - & - & - & - & - & 84.5 & - \\
Dependency Tree-LSTM  &  - & - & - & - & - & - & - & 0.868 & - & - \\
\hline
\end{tabular}
}
\caption{{\bf Transfer test results for various architectures trained in different ways}. Underlined are best results for transfer learning approaches, in bold are best results among the models trained in the same way. $^{\dagger}$ indicates methods that we trained, other transfer models have been extracted from \cite{hill2016learning}. For best published supervised methods (no transfer), we consider AdaSent \cite{zhao2015self}, TF-KLD \cite{ji2013discriminative}, Tree-LSTM \cite{tai2015improved} and Illinois-LH system \cite{lai2014illinois}. (*) Our model trained on SST obtained 83.4 for MR and 86.0 for SST (MR and SST come from the same source), which we do not put in the tables for fair comparison with transfer methods.
\label{table:tasks_results}
}
\end{table*}

\begin{figure}[h!]
  \centering
  \includegraphics[width=1.\linewidth]{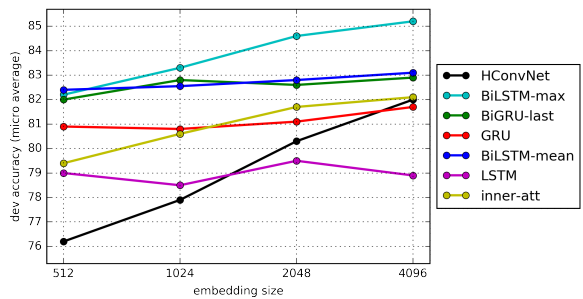}
    \caption{{\bf Transfer performance w.r.t. embedding size} using the micro aggregation method.
    \label{fig:embed_results}}
\end{figure}

\section{Empirical results}
In this section, we refer to "micro" and "macro" averages of development set (dev) results on transfer tasks whose metrics is accuracy: we compute a "macro" aggregated score that corresponds to the classical average of dev accuracies, and the "micro" score that is a sum of the dev accuracies, weighted by the number of dev samples.

\subsection{Architecture impact}

\paragraph{Model}
We observe in Table~\ref{table:archi_results} that different models trained on the same NLI corpus lead to different transfer tasks results. The BiLSTM-4096 with the max-pooling operation performs best on both SNLI and transfer tasks. Looking at the micro and macro averages, we see that it performs significantly better than the other models LSTM, GRU, BiGRU-last, BiLSTM-Mean, inner-attention and the hierarchical-ConvNet.

Table~\ref{table:archi_results} also shows that better performance on the training task does not necessarily translate in better results on the transfer tasks like when comparing inner-attention and BiLSTM-Mean for instance.

We hypothesize that some models are likely to over-specialize and adapt too well to the biases of a dataset without capturing general-purpose information of the input sentence. For example, the inner-attention model has the ability to focus only on certain parts of a sentence that are useful for the SNLI task, but not necessarily for the transfer tasks. On the other hand, BiLSTM-Mean does not make sharp choices on which part of the sentence is more important than others. The difference between the results seems to come from the different abilities of the models to incorporate general information while not focusing too much on specific features useful for the task at hand.

For a given model, the transfer quality is also sensitive to the optimization algorithm: when training with Adam instead of SGD, we observed that the BiLSTM-max converged faster on SNLI (5 epochs instead of 10), but obtained worse results on the transfer tasks, most likely because of the model and classifier's increased capability to over-specialize on the training task.

\begin{table*}[t]
 \begin{center}
 \begin{small}
  \begin{tabular}{l@{\,}l|cccc|cccc}
    & &
      \multicolumn{4}{c}{\bf Caption Retrieval} &
      \multicolumn{4}{c}{\bf Image Retrieval} \\
     \multicolumn{2}{l}{\bf Model} & \bf R@1 &\bf R@5 & \bf R@10 & \bf Med r & \bf R@1 & \bf R@5 & \bf R@10 & \bf Med r \\
\hline
\hline
\multicolumn{7}{l}{\it Direct supervision of sentence representations} && \\
\hline        
    \textit{m}-CNN & \cite{ma2015multimodal} & 38.3 & - & 81.0 & 2 & 27.4 & - & 79.5 & 3 \\
    \textit{m}-CNN$_{\text{ENS}}$ & \cite{ma2015multimodal}& 42.8 & - & 84.1 & 2 & 32.6 & - & 82.8 & 3 \\
    Order-embeddings & \cite{vendrov2015order}& \bf 46.7 & - & \bf 88.9 & 2 & \bf 37.9 & - & \bf 85.9 & 2 \\
\hline
\hline
\multicolumn{7}{l}{\it Pre-trained sentence representations} && \\
\hline    
    SkipThought & + VGG19 (82k) & 33.8 & 67.7 & 82.1 & 3 & 25.9 & 60.0 & 74.6 & 4 \\
    SkipThought & + ResNet101 (113k)  & 37.9 & 72.2 & 84.3 & 2	& 30.6 & 66.2 & 81.0 & 3 \\
    BiLSTM-Max (on SNLI) & + ResNet101 (113k) & 42.4 & \bf 76.1 & 87.0 & 2 & 33.2 & \bf 69.7 & 83.6 & 3 \\
    BiLSTM-Max (on AllNLI) & + ResNet101 (113k) & \bf 42.6 &  75.3 & \bf 87.3 & 2 & \bf 33.9 & \bf 69.7 & \bf 83.8 & 3 \\
  \end{tabular}
  \end{small}
\caption{\label{table:coco_results} {\bf COCO retrieval results}. SkipThought is trained either using 82k training samples with VGG19 features, or with 113k samples and ResNet-101 features (our setting). We report the average results on 5 splits of 1k test images.}

\end{center}
\end{table*}

\paragraph{Embedding size}
Figure~\ref{fig:embed_results} compares the overall performance of different architectures, showing the evolution of micro averaged performance with regard to the embedding size. 

Since it is easier to linearly separate in high dimension, especially with logistic regression, it is not surprising that increased embedding sizes lead to increased performance for almost all models. However, this is particularly true for some models (BiLSTM-Max, HConvNet, inner-att), which demonstrate unequal abilities to incorporate more information as the size grows. We hypothesize that such networks are able to incorporate information that is not directly relevant to the objective task (results on SNLI are relatively stable with regard to embedding size) but that can nevertheless be useful as features for transfer tasks.

\subsection{Task transfer}
We report in Table~\ref{table:tasks_results} transfer tasks results for different architectures trained in different ways. We group models by the nature of the data on which they were trained. The first group corresponds to models trained with unsupervised unordered sentences. This includes bag-of-words models such as word2vec-SkipGram, the Unigram-TFIDF model, the Paragraph Vector model \cite{le2014distributed}, the Sequential Denoising Auto-Encoder (SDAE) \cite{hill2016learning} and the SIF model \cite{arora2016asimple}, all trained on the Toronto book corpus \cite{zhu2015aligning}. The second group consists of models trained with unsupervised ordered sentences such as FastSent and SkipThought (also trained on the Toronto book corpus). We also include the FastSent variant ``FastSent+AE'' and the SkipThought-LN version that uses layer normalization. We report results from models trained on supervised data in the third group, and also report some results of supervised methods trained directly on each task for comparison with transfer learning approaches.

\paragraph{Comparison with SkipThought}
The best performing sentence encoder to date is the SkipThought-LN model, which was trained on a very large corpora of ordered sentences. With much less data (570k compared to 64M sentences) but with high-quality supervision from the SNLI dataset, we are able to consistently outperform the results obtained by SkipThought vectors. We train our model in less than a day on a single GPU compared to the best SkipThought-LN network trained for a month. Our BiLSTM-max trained on SNLI performs much better than released SkipThought vectors on MR, CR, MPQA, SST, MRPC-accuracy, SICK-R, SICK-E and STS14 (see Table~\ref{table:tasks_results}). Except for the SUBJ dataset, it also performs better than SkipThought-LN on MR, CR and MPQA. We also observe by looking at the STS14 results that the cosine metrics in our embedding space is much more semantically informative than in SkipThought embedding space (pearson score of 0.68 compared to 0.29 and 0.44 for ST and ST-LN). We hypothesize that this is namely linked to the matching method of SNLI models which incorporates a notion of distance (element-wise product and absolute difference) during training.

\paragraph{NLI as a supervised training set}
Our findings indicate that our model trained on SNLI obtains much better overall results than models trained on other supervised tasks such as COCO, dictionary definitions, NMT, PPDB \cite{ganitkevitch2013ppdb} and SST. 
For SST, we tried exactly the same models as for SNLI; it is worth noting that SST is smaller than NLI.
Our representations constitute higher-quality features for both classification and similarity tasks. One explanation is that the natural language inference task constrains the model to encode the semantic information of the input sentence, and that the information required to perform NLI is generally discriminative and informative.

\paragraph{Domain adaptation on SICK tasks}
Our transfer learning approach obtains better results than previous state-of-the-art on the SICK task - can be seen as an out-domain version of SNLI - for both entailment and relatedness. We obtain a pearson score of 0.885 on SICK-R while \cite{tai2015improved} obtained 0.868, and we obtain 86.3\% test accuracy on SICK-E while previous best hand-engineered models \cite{lai2014illinois} obtained 84.5\%.
We also significantly outperformed previous transfer learning approaches on SICK-E \cite{bowman2015large} that used the parameters of an LSTM model trained on SNLI to fine-tune on SICK (80.8\% accuracy). We hypothesize that our embeddings already contain the information learned from the in-domain task, and that learning only the classifier limits the number of parameters learned on the small out-domain task.

\paragraph{Image-caption retrieval results}
In Table~\ref{table:coco_results}, we report results for the COCO image-caption retrieval task. We report the mean recalls of 5 random splits of 1K test images. When trained with ResNet features and 30k more training data, the SkipThought vectors perform significantly better than the original setting, going from 33.8 to 37.9 for caption retrieval R@1, and from 25.9 to 30.6 on image retrieval R@1. Our approach pushes the results even further, from 37.9 to 42.4 on caption retrieval, and 30.6 to 33.2 on image retrieval. These results are comparable to previous approach of \cite{ma2015multimodal} that did not do transfer but directly learned the sentence encoding on the image-caption retrieval task. This supports the claim that pre-trained representations such as ResNet image features and our sentence embeddings can achieve competitive results compared to features learned directly on the objective task.

\paragraph{MultiGenre NLI} The MultiNLI corpus
\cite{williams2017broad} was recently released as a multi-genre version of SNLI. With 433K sentence pairs, MultiNLI improves upon SNLI in its coverage: it contains ten distinct genres of written and spoken English, covering most of the complexity of the language. We augment Table 4 with our model trained on both SNLI and MultiNLI (AllNLI). We observe a significant boost in performance overall compared to the model trained only on SLNI. Our model even reaches AdaSent performance on CR, suggesting that having a larger coverage for the training task helps learn even better general representations. On semantic textual similarity STS14, we are also competitive with PPDB based paragram-phrase embeddings with a pearson score of 0.70. Interestingly, on caption-related transfer tasks such as the COCO image caption retrieval task, training our sentence encoder on other genres from MultiNLI does not degrade the performance compared to the model trained only SNLI (which contains mostly captions), which confirms the generalization power of our embeddings.

\section{Conclusion}
This paper studies the effects of training sentence embeddings with supervised data by testing on 12 different transfer tasks. We showed that models learned on NLI can perform better than models trained in unsupervised conditions or on other supervised tasks. By exploring various architectures, we showed that a BiLSTM network with max pooling makes the best current universal sentence encoding methods, outperforming existing approaches like SkipThought vectors.

We believe that this work only scratches the surface of possible combinations of models and tasks for learning generic sentence embeddings. Larger datasets that rely on natural language understanding for sentences could bring sentence embedding quality to the next level.

\bibliography{emnlp2017}
\bibliographystyle{emnlp_natbib}

\clearpage
\newpage

\section*{Appendix}
\paragraph{Max-pooling visualization for BiLSTM-max trained and untrained}

Our representations were trained to focus on parts of a sentence such that a classifier can easily tell the difference between contradictory, neutral or entailed sentences.

In Table~\ref{fig:qualitative_trained} and Table~\ref{fig:qualitative_untrained}, we investigate how the max-pooling operation selects the information from the hidden states of the BiLSTM, for our trained and untrained BiLSTM-max models (for both models, word embeddings are initialized with GloVe vectors).

For each time step $t$, we report the number of times the max-pooling operation selected the hidden state $h_t$ (which can be seen as a sentence representation centered around word $w_t$). 

Without any training, the max-pooling is rather even across hidden states, although it seems to focus consistently more on the first and last hidden states. When trained, the model learns to focus on specific words that carry most of the meaning of the sentence without any explicit attention mechanism.

Note that each hidden state also incorporates information from the sentence at different levels, explaining why the trained model also incorporates information from all hidden states.

\begin{figure}[h!]
  \centering
  \includegraphics[width=0.9\linewidth]{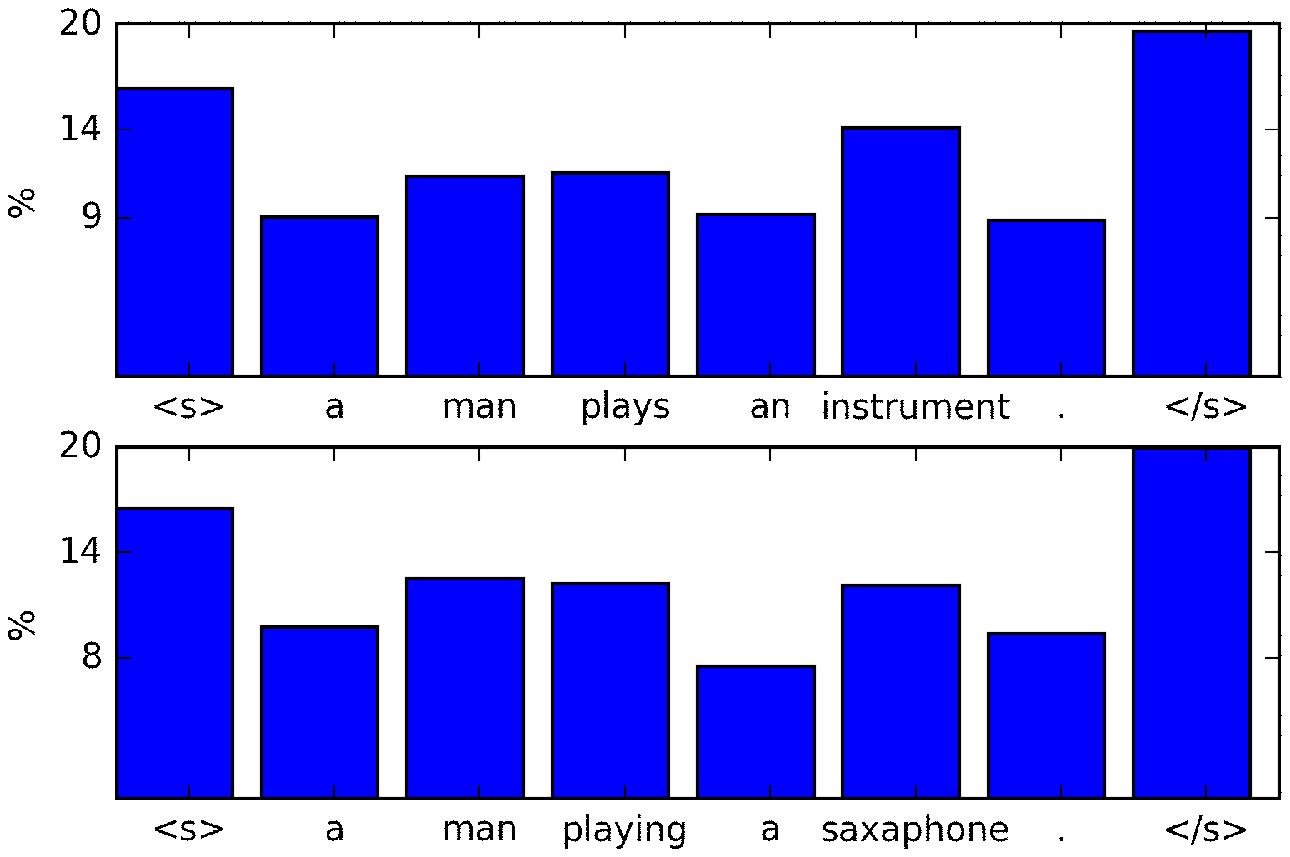}
    \caption{\label{fig:qualitative_trained} Pair of entailed sentences A: Visualization of max-pooling for BiLSTM-max 4096 \textbf{untrained}.}
\end{figure}
\begin{figure}[h!]
  \centering
  \includegraphics[width=1.0\linewidth]{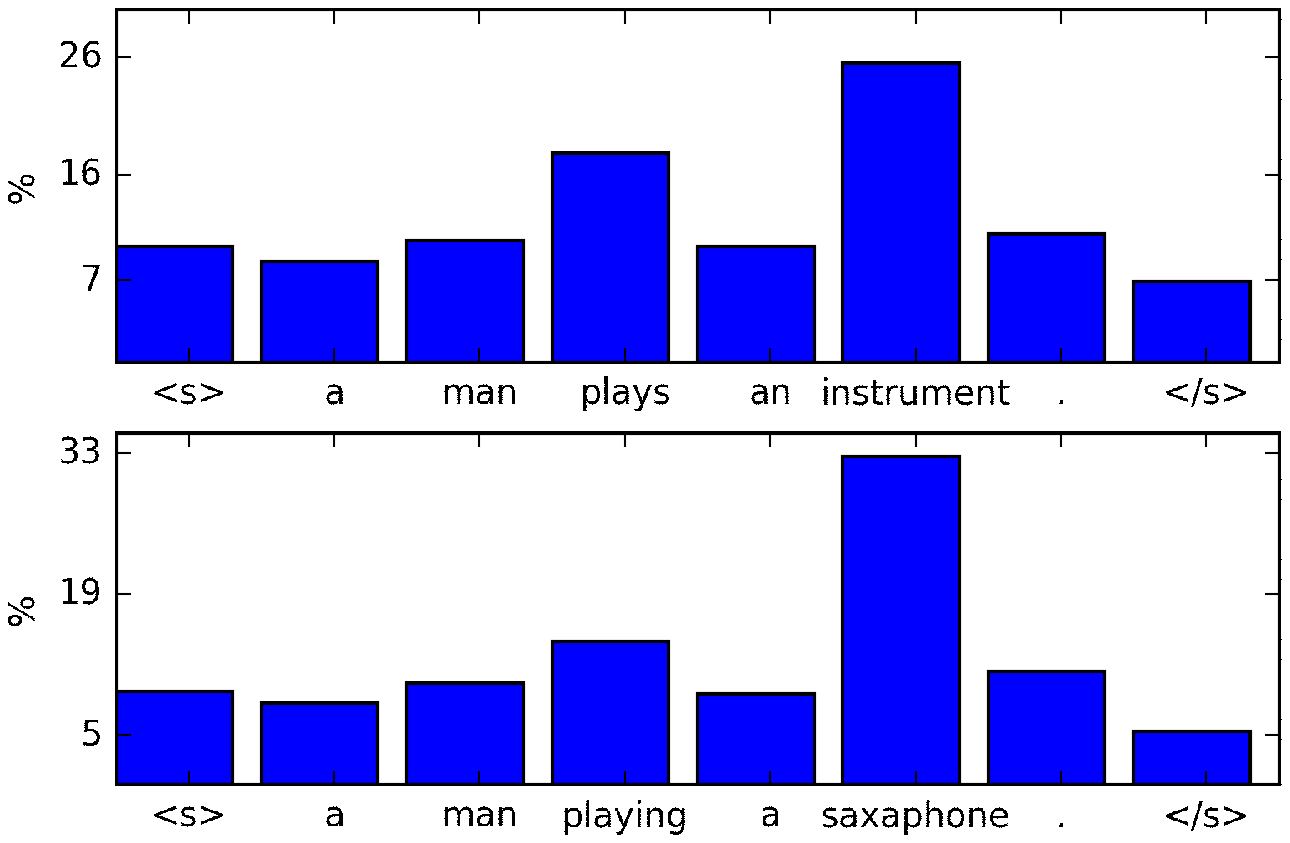}
    \caption{\label{fig:qualitative_untrained}Pair of entailed sentences A: Visualization of max-pooling for BiLSTM-max 4096 \textbf{trained on NLI}.}
\end{figure}

\begin{figure}[h!]
  \centering
  \includegraphics[width=1.0\linewidth]{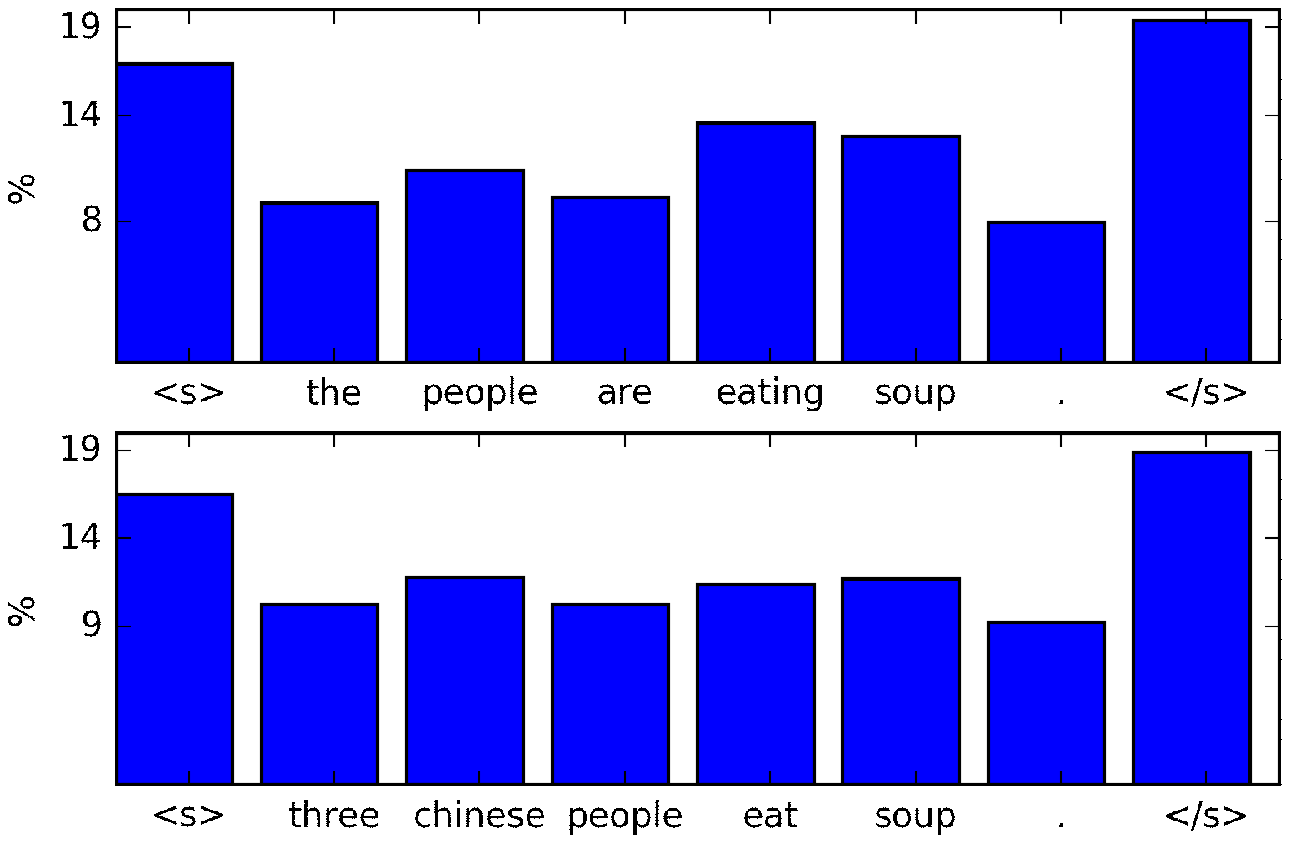}
    \caption{\label{fig:qualitative_trained} Pair of entailed sentences B: Visualization of max-pooling for BiLSTM-max 4096 \textbf{untrained}.}
\end{figure}

\begin{figure}[h!]
  \centering
  \includegraphics[width=1.0\linewidth]{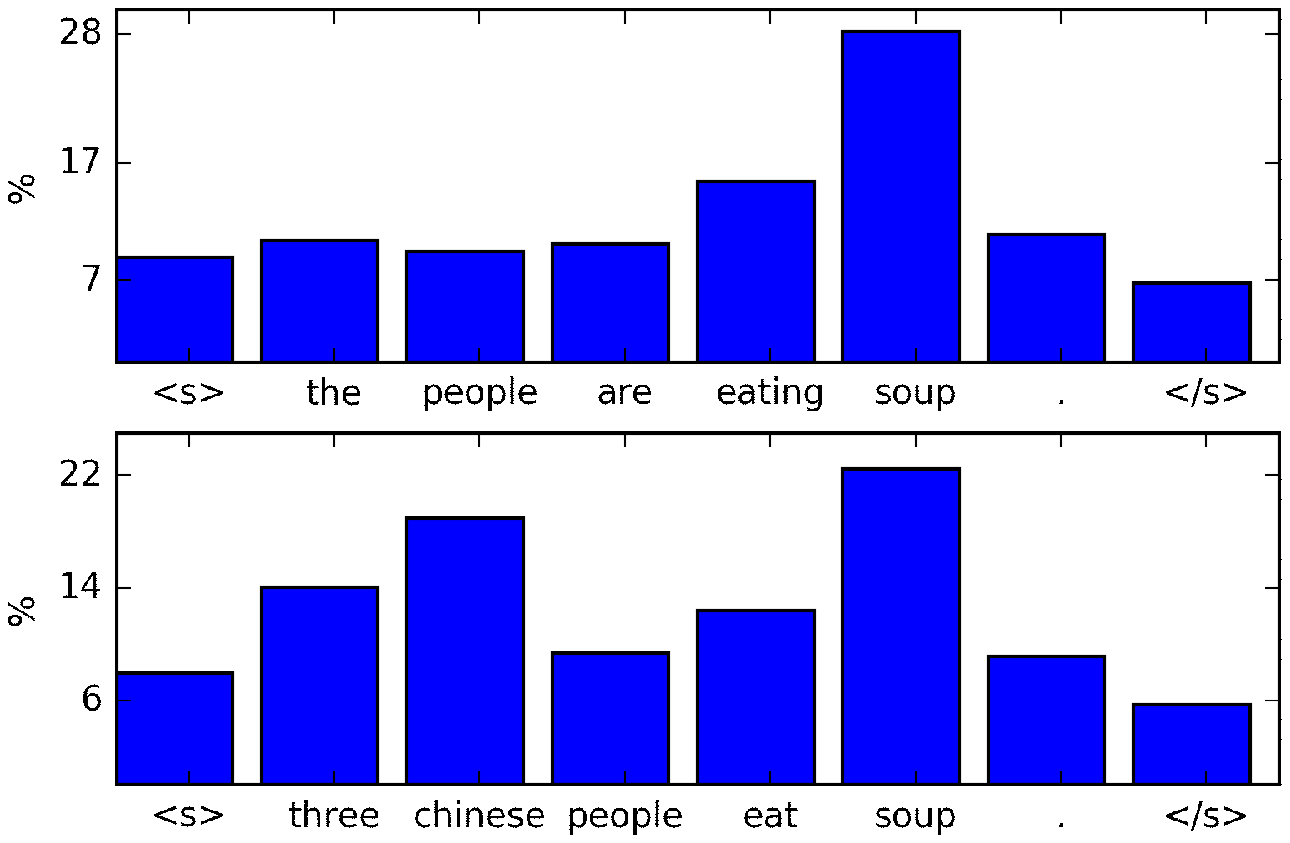}
    \caption{\label{fig:qualitative_untrained}Pair of entailed sentences B: Visualization of max-pooling for BiLSTM-max 4096 \textbf{trained on NLI}.}
\end{figure}

\end{document}